\documentclass[letterpaper, 10 pt, conference]{ieeeconf}  

\IEEEoverridecommandlockouts                              

\usepackage{amsmath}
\usepackage{amssymb}
\usepackage{hyperref}
\usepackage{url}
\usepackage{graphics}
\usepackage{graphicx}
\usepackage{xcolor}         
\usepackage{booktabs}       

\usepackage{wrapfig}
\usepackage{comment}
\usepackage{caption}

\usepackage{multirow}

\usepackage[ruled,vlined]{algorithm2e}

\usepackage{amsmath}

\newcommand{\rowsqueeze}{\vspace{-2pt}}

\title{\LARGE \bf
Embodied Instruction Following in Unknown Environments
}

\author{Zhenyu Wu$^{1}$, Ziwei Wang$^{2}$, Xiuwei Xu$^{3}$, Hang Yin$^{3}$, Yinan Liang$^{3}$, Angyuan Ma$^{3}$, Jiwen Lu$^{3}$ and Haibin Yan$^{*1}$
\thanks{*Corresponding author.}
\thanks{$^{1}$Zhenyu Wu and Haibin Yan are with the School of Intelligent Engineering and
Automation, Beijing University of Posts and Telecommunications. {\tt\small \{wuzhenyu, eyanhaibin\}@bupt.edu.cn.}}
\thanks{$^{3}$Ziwei Wang is with the School of Electrical and Electronic Engineering, Nanyang Technological University. {\tt\small \{ziwei.wang\}@ntu.edu.sg}}
\thanks{$^{3}$Xiuwei Xu, Hang Yin, Yinan Liang, Angyuan Ma, and Jiwen Lu are with the Department of Automation, Tsinghua University.  {\tt\small \{xxw21, yinh23,liangyn24,maay21\}@mails.tsinghua.edu.cn.} {\tt\small lujiwen@tsinghua.edu.cn.}}
}

\begin{document}

\maketitle
\thispagestyle{empty}
\pagestyle{empty}

\begin{abstract}

Enabling embodied agents to complete complex human instructions from natural language is crucial to autonomous systems in household services. Conventional methods can only accomplish human instructions in the known environment where all interactive objects are provided to the embodied agent, and directly deploying the existing approaches for the unknown environment usually generates infeasible plans that manipulate non-existing objects. On the contrary, we propose an embodied instruction following (EIF) method for complex tasks in the unknown environment, where the agent efficiently explores the unknown environment to generate feasible plans with existing objects to accomplish abstract instructions. Specifically, we build a hierarchical embodied instruction following framework including the high-level task planner and the low-level exploration controller with multimodal large language models. We then construct a semantic representation map of the scene with dynamic region attention to demonstrate the known visual clues, where the goal of task planning and scene exploration is aligned for human instruction. For the task planner, we generate the feasible step-by-step plans for human goal accomplishment according to the task completion process and the known visual clues. For the exploration controller, the optimal navigation or object interaction policy is predicted based on the generated step-wise plans and the known visual clues. 
The experimental results demonstrate that our method can achieve 45.09\% success rate in 204 complex human instructions such as making breakfast and tidying rooms in large house-level scenes. Code and supplementary are available at \href{https://gary3410.github.io/eif_unknown/}{https://gary3410.github.io/eif\_unknown/}.

\end{abstract}

\section{INTRODUCTION}

Building intelligent autonomous systems \cite{huang2023voxposer, mu2024embodiedgpt, yin2024sg, kumar2024open} to complete household tasks such as making breakfast and tidying rooms is highly demanded to reduce the laborer cost in our daily life. The agent is required to understand the visual clues of the surrounding scene and the language instructions, and feasible action plans are then generated for object interaction with the goal of high success rate and low action cost to accomplish human demands.

\begin{figure*}[t]
  \centering
 \includegraphics[width=1.0\textwidth]{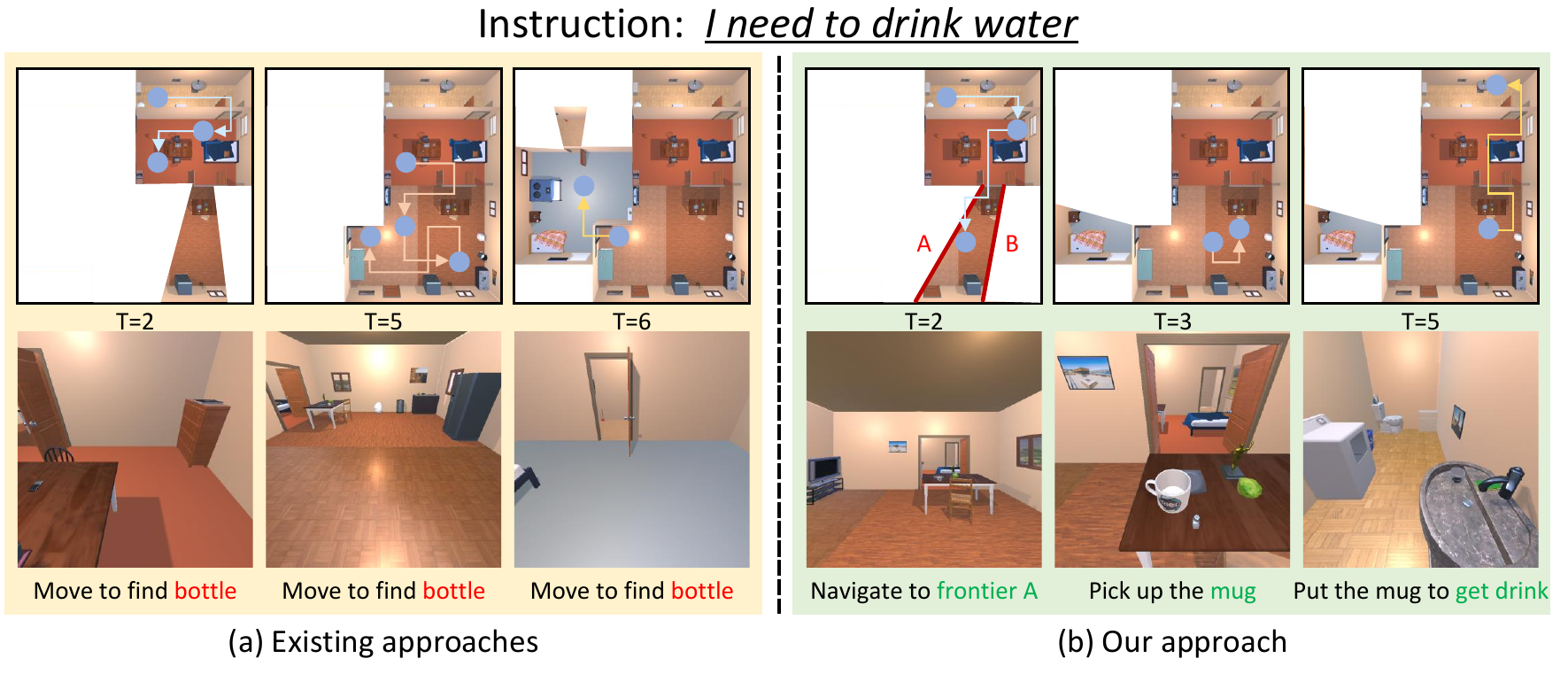}
  \caption{Comparison between conventional EIF methods and our approach in unknown environments. Existing methods fail to complete the instruction even with long exploration cost, while our method efficiently achieves the goal with efficient navigation and object interaction.}
  \label{fig_1: comparison}
  \vspace{-0.6cm}
\end{figure*}

To achieve this, end-to-end methods \cite{pashevich2021episodic} directly generate the low-level actions from raw image input and natural language with the supervision of expert trajectories. To reduce the learning difficulties in the complex task, modular methods \cite{ding2023embodied},  sequentially learn the instruction comprehension, state perception, spatial memory construction, high-level planning and low-level control to complete human goals. Since embodied agents are expected to complete more diverse and complex instructions, large language models (LLMs) are widely employed in EIF \cite{wu2023embodied} due to their strong reasoning power and high generalization ability. However, existing methods can only generate plans in known environments where categories of all interactable objects in the scene are given to LLMs. Since the agent does not know the objects in the unknown environment, the generated plans are usually infeasible because of interacting with non-existing objects. 
Fig. \ref{fig_1: comparison} (a) demonstrates an example of existing methods, where the agent is unaware that no bottles exist in the unknown environment. Interacting with the non-existent bottles based on the infeasible plan fails to accomplish the human goals of water serving.

In realistic deployment scenarios, household agents usually work in unknown environments without stored scene maps. Building scene maps in advance cannot accurately represent the scene, where object properties such as location and existence change frequently due to human activity in daily life. For example, the mug may be on the dining table and the coffee table respectively when humans are having dinner and watching TV. Meanwhile, potatoes might have been consumed and tomatoes are then purchased for the next breakfast. Therefore, failing to generate feasible plans in unknown environments strictly limits the practicality of the embodied agents. The agent working in realistic deployment scenarios is required to build real-time scene maps, where feasible plans are generated with minimal exploration cost.

In this paper, we propose an EIF method for complex tasks in the unknown environment. Different from conventional methods that assume knowing interactable objects in advance, our method navigates the unknown environment to efficiently discover objects that are relevant to the complex human requirements. Therefore, the embodied agent can generate feasible task plans in realistic indoor scenes where the locations and existence of objects are frequently changing.
Fig. \ref{fig_1: comparison} (b) also demonstrates the same example of water serving implemented by our method, and our agent efficiently discovers the mug and uses it as the receptacle of water because no bottles exist in the scene.
We first construct a hierarchical EIF framework including the high-level task planner and the low-level exploration controller with multi-modal LLMs, which are finetuned by the large-scale generated trajectories of the complex EIF tasks. We then design a scene-level semantic representation map to depict the visual clues in the known area, through which the goals of the task planner and the exploration controller can be aligned to feasibly complete human instructions.

More specifically, the goal of the task planner is to generate feasible plans for human instruction including navigation and manipulation in natural language. The task planner predicts the next step based on the semantic representation map and the task completion process. The exploration controller aims at discovering task-related objects with low action cost, which selects the optimal navigation policy from all navigable borders or object interaction policy according to the semantic representation map and the generated step-wise plans. For the scene-level semantic feature map, we project the CLIP features of collected RGB images during exploration to the top-down map with dynamic region attention, which preserves the task-relevant visual information in the map without redundancy. 
The experimental results in ProcTHOR \cite{deitke2022️} simulation environment show that our method can achieve 45.09\% success rate in 204 complex human instructions in large house-level scenes.

\begin{figure*}[t]
  \centering
  \includegraphics[width=1.0\textwidth]{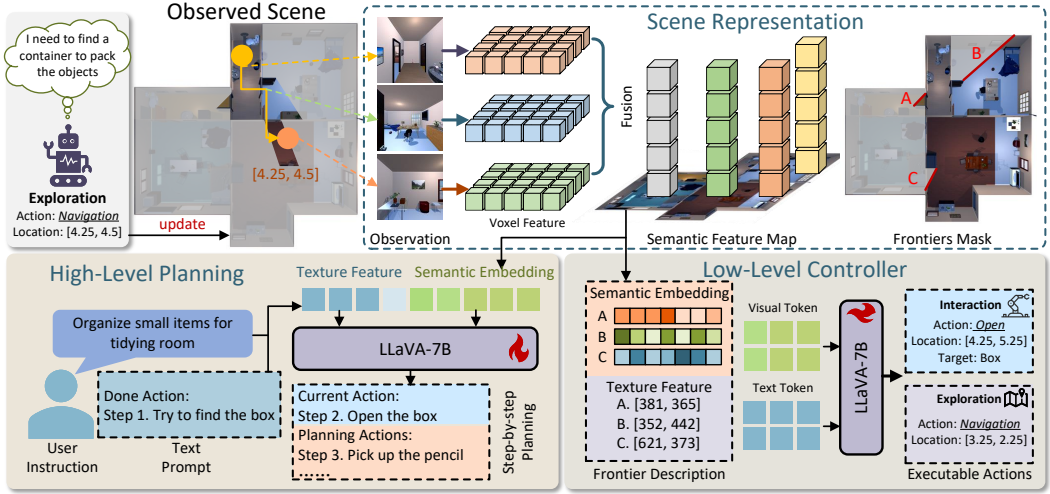}
  \caption{Overview of our approach. The scene feature map is constructed based on real-time RGB-D images, which is leveraged as visual clues for the high-level planner and the low-level controller. The planner generates the step-wise plans, which are leveraged to predict the specific actions in the controller. The optimal border between unknown and known regions is selected for scene exploration, and the scene feature map is updated with the visual clues seen in during the exploration.}
  \vspace{-0.6cm}
  \label{fig_2:pipeline}
\end{figure*}

\section{Related Works}
\textbf{Embodied Instruction Following:}
The EIF task requires the robot to follow human instructions represented by natural language in the interactive environment.  
A key challenge for the EIF task is generating interaction goals and actions grounded in the deployment environment according to the instructions.
Prior works (e.g., LACMA \cite{yang2023lacma}, E.T. \cite{pashevich2021episodic}, M-TRACK \cite{song2022one}) have explored end-to-end transformer architecture to generate grounded low-level interaction actions based on the current environment perception, modular approaches (e.g., HLSM \cite{blukis2022persistent}, FILM \cite{min2021film}, LLM-Planner \cite{song2023llm}) propose enhancing the generalization of unseen scenes with hierarchical planners.
However, prior arts have focused on single-room environments, which are designed for known environments where visual clues of the whole scene can be easily acquired by looking around.
The low scalability of the scene scale limits their ability to discover required visual clues in unknown environments for feasible action generation.

\textbf{Scene Representation for Visual-language Navigation:}
Existing scene representations consist of three categories: 2D semantic maps, 3D geometric maps and scene graphs. Early works \cite{batra2020objectnav} constructed the 2D semantic maps by projecting visual clues in the top-down view, which are leveraged for navigation frontier selection for target finding. L3MVN \cite{yu2023l3mvn} determined the semantic relevance of the objects around each frontier to the target by BERT \cite{devlin2018bert}. 
To embed the geometric information, 3D geometric maps are investigated by fusing the structure and semantic information. 
ConceptFusion \cite{jatavallabhula2023conceptfusion} integrated fine-grained alignment of semantic features with 3D maps in SLAM, multi-view fusion, and NeRF \cite{mildenhall2021nerf} for multiple downstream tasks.
To reduce the storage overhead, scene graphs \cite{rana2023sayplan, yin2025unigoal} are proposed to represent objects or concepts as nodes and spatial relations as edges to represent the scene topology efficiently.
Inspired by the above approaches, we construct semantic feature maps to empower embodied agents to explore unknown environments.

\section{Problem Statement}
Given the human instruction $I$ in natural language, the robot should generate a sequence of action primitives including (\texttt{PickUp, Place, Open, Close, ToggleOn, ToggleOff, Slice}) to complete the instruction.
The agent can only acquire the scene information for instruction following via an RGB-D camera mounted on the agent, through which the agents build a semantic map $S$ to generate the feasible interaction.
In realistic deployment, the embodied agent usually works in unknown environments, where the location and existence of objects in the house-level scene are not known. Therefore, we add an additional action primitive (\texttt{Navigation}) to enable the agent to explore the scene for visual information collection.

The agent consists of a high-level planner that reasons step-by-step plans $P=\{p_i\}_{i=1}^{T}$ from human instructions and a low-level controller that predicts the specific actions $A=\{a_j^i\}_{j=1}^{\tau_i}$ for each step for scene navigation or object interaction. $T$ means the number of steps to achieve the human goal, and $\tau_i$ is the number of special actions to achieve the $i_{th}$ step in the high-level plan. The high-level planner is represented by natural language (e.g. Step 2. Heat the potato) given the human instruction (e.g. Can you make breakfast for me?), and the low-level controller transfers the step-by-step plans into executable actions with action primitives, location and target objects (e.g. \texttt{Place}, potato, (10, 8) or \texttt{Navigate}, frontier, (2, 3)). Finally, the agent only manipulates the existing relevant objects to achieve human goals.

\section{Approach}

\subsection{Overall Pipeline}
In realistic deployment scenarios of household robots, the physical world is usually unknown for the agent because the existence and locations frequently change due to human activity.  
Therefore, the agents are required to construct the online scene feature map according to the real-time visual perception during the robot navigation, through which the agent generates feasible step-by-step plans to achieve the human goal and the efficient exploration trajectories for the unknown scene including navigation and object interaction to complete each step in the plan.
Fig. \ref{fig_2:pipeline} demonstrates the overall pipeline of our agent. The scene feature map represents the visual clues of the scene in the top-down view based on the collected RGB-D images during exploration, where the pre-trained features of regions with higher relevance to the instruction are assigned higher importance for feature map construction. 
The high-level planner generates the plans for the next step with natural language based on the task completion process and the semantic feature map, and the low-level controller predicts the templated action primitives, location and target objects for executable navigation or manipulation based on the scene feature map and the plan for the next step.

\begin{figure*}[t]
  \centering
  \includegraphics[width=1.0\textwidth]{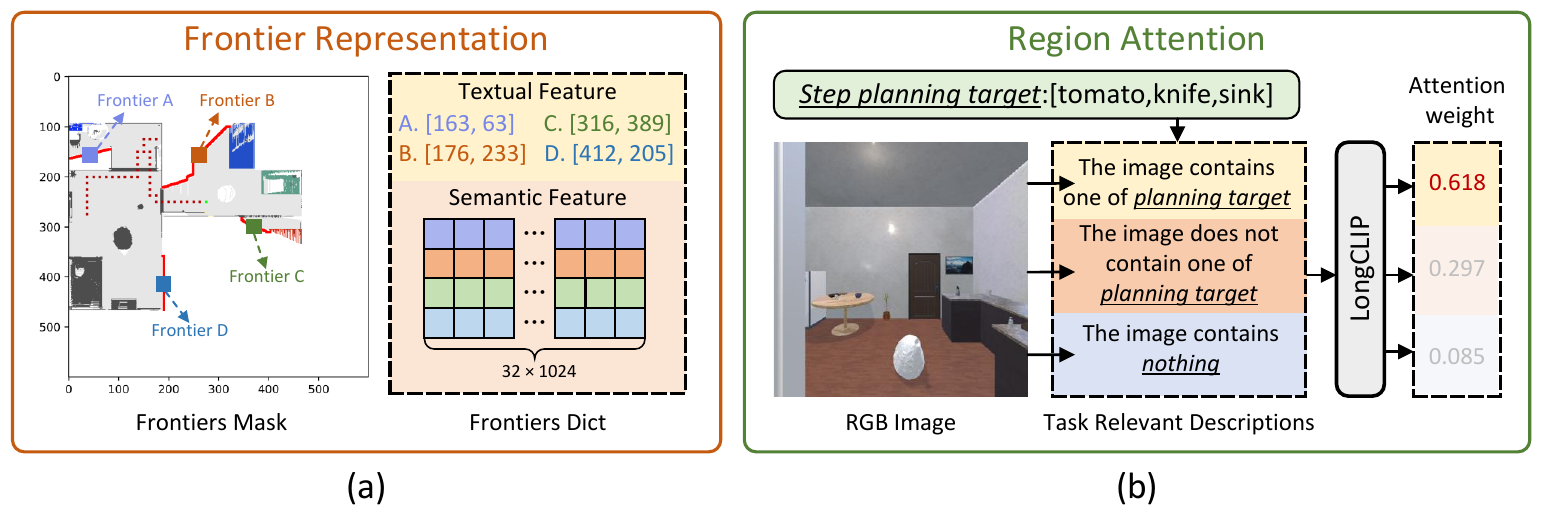}
  \caption{Details of frontier representation and region attention weights.}
  \label{fig:frontier}
  \vspace{-0.6cm}
\end{figure*}

\subsection{Hierarchical Agents for EIF in Unknown Environments}


\textbf{High-level planner:}
Formally, the high-level planner $P^{h}$ decomposes human instructions into step by step, generating the plan for the next step in natural language, which considers textual information including human instruction and completed steps and the visual clues represented by semantic feature maps.
\begin{align}
    p_i = P^{h}(I, \{p_k\}_{k=1}^{i-1}; S_{i-1})
\end{align}
where $S_{i}$ means the semantic feature maps updated in the $i_{th}$ step.

\textbf{Low-level controller:}
The low-level controller predicts $P^{l}$ the specific actions, including action primitives, locations, and target objects according to the high-level plans generated and the semantic feature maps, which explore the unknown scene and complete the stepwise plan.
\begin{align}
\{a_j^i, l_j^i, o_j^i\}=P^{l}(p_i, \{f_{m}^{i}\}_m; \{s_{m}^{i}\}_m)
\end{align}
where $l_j^i$ and $o_j^i$ are the predicted location and target objects for the $j_{th}$ actions in the $i_{th}$ step of the high-level plan. Meanwhile, \smash{$f_{m}^{i}$} means the textual features of the $m_{th}$ segment of the frontier between known and unknown regions for $S_{i}$, where $m$ represents the number of frontier segments in the entire $S_{i}$.
The textual features are demonstrated by the coordinates of the middle point for the frontier segment. $s_{m}^{i}$ denotes the semantic features of the $m_{th}$ frontier segments, which is demonstrated by the semantic feature map patches containing the corresponding frontiers. 
The low-level controller not only explores the unknown scene with navigation and object interaction but also completes the step-wise plans by manipulating the target object (e.g. pick up the tomato). 
For action primitives except for \texttt{navigate}, the predicted actions are implemented on the target objects. For \texttt{navigate}, the robot just moves to the predicted locations without object interaction.

Both high-level $P^{h}$ and low-level $P^l$ are composed with VLMs, which leverage scene semantic information to decompose instructions into step-by-step plans and ground them into executable low-level actions.

\subsection{Online Semantic Feature Maps}
The high-level planner and the low-level controller should be aligned so that they can generate feasible plans and exploratory actions to achieve human instructions in an unknown environment.
In realistic deployment scenarios of household robots, the existence and locations frequently change due to human activity. Therefore, we propose an online semantic feature map that is dynamically updated during the exploration of the unknown scene for each human instruction.

\textbf{Feature extraction: }To enable semantic feature maps to acquire high generalization ability in various human instructions, we use vision foundation model to extract pixelwise visual features $\mathbf{f}_{xy}^i$ at time $i$ for the pixel in $x_{th}$ row and $y_{th}$ column of the RGB image $C_{i}$ by combining the feature of the entire image and that of the instance mask containing the corresponding pixel. 
The visual features contribute to the projected location in the scene feature map via the depth image $D_{i}$.
\begin{align}
\mathbf{F}_{uv}^{i}=\sum_{x,y}\mathbf{f}_{xy}^i\cdot\mathbb{I}(\mathcal{P}((x,y), D_i)\in \mathcal{S}(u,v))
\end{align}
where $\mathbf{F}_{uv}^{i}$ means the contribution to the element in the $u_{th}$ row and $v_{th}$ column of the semantic feature map from the visual information collected in time $i$, and $\mathcal{P}$ demonstrates the projected function. $\mathcal{S}(u,v)$ means the pixel in the $u_{th}$ row and $v_{th}$ column in the semantic feature map, and the indicator function $\mathbb{I}(\cdot)$ equals one for true and zero otherwise. 

\textbf{Frontier representation:}
We generate frontier masks that distinguish between known and unknown regions based on the occupancy map for efficient exploration.
Through connected component analysis, we obtain the mask of each frontier instance.
We further remove frontiers with areas smaller than the threshold (150 pixels) to reduce redundant exploration.
We sample 32 visual embeddings as frontier tokens according to the frontier instance mask on the corresponding region of the feature map, while utilizing the coordinates of their centroids for the frontier text description. The specific representation is illustrated in Fig. \ref{fig:frontier} (a).

\textbf{Dynamic region attention: }Since the house for embodied instruction following in realistic world is usually very large, regarding all images with equal importance in semantic feature map construction leads to significant information redundancy. Meanwhile, different visual clues usually make various contribution to the given human instruction.
Therefore, we should assign large importance to relevant visual clues when updating the semantic feature maps with task demanded objects $L_{dec}$, so that sufficient visual information can be represented without redundancy for high-level planning and low-level exploration.
Specifically, we further expand $L_{dec}$ into $\overline{L_{dec}}$ and $L_{none}$ to match the input requirements of image and text alignment models such as CLIP. $\overline{L_{dec}}$ and $L_{none}$ describe the image as not containing the target objects and not containing the objects, respectively.
The components as illustrated in Fig. \ref{fig:frontier} (b), and consider the score of $L_{dec}$ as the attention score $c_i$.
Finally, the online semantic feature map is updated with dynamic region attention:
\begin{align}
    \mathbf{S}_{uv}^{i}=(1-w_i) \mathbf{S}_{uv}^{i-1}+w_i \mathbf{F}_{uv}^{i}, \quad w_i=c_{i}/\frac{1}{i} \sum_{k=1}^{i}c_{k}
\end{align}
where $\mathbf{S}_{uv}^{i}$ means the features in the $i_{th}$ row and $j_{th}$ column of the semantic feature maps at time $i$. 
The normalized weight $w_i$ represents the importance of the current semantic features compared with known visual clues, where $c_{k}$ is the original similarity score between the image and the prompt in the $k_{th}$ time step. 

\begin{table*}[t]
\caption{Comparison with different EIF methods across different instructions in the ProcTHOR simulator, where LLM-P$^{*}$ represents the LLM-P without performing re-planning.}
\centering
\vspace{-2mm}
\footnotesize
\setlength{\tabcolsep}{12pt}
\begin{tabular}{@{}lcccccccccc@{}}
\toprule
\multirow{2}{*}{\textbf{Method}}& \multicolumn{5}{c}{\textbf{Normal-scale}}
& \multicolumn{5}{c}{\textbf{Large-scale}} \rowsqueeze \\
 \cmidrule{2-6}\cmidrule{6-11} \rowsqueeze
 & SR & PLWSR & GC & PLWGC & Path & SR & PLWSR & GC & PLWGC & Path \\
 \midrule
 \multicolumn{11}{l}{\textbf{Target-specific Short}} \rowsqueeze\\
 \midrule
 LLM-P$^{*}$ & 27.86 & 23.49 & 41.50 & 35.35 & 25.27 & 17.16 & 11.70 & 33.25 & 22.87 & 65.75 \\
 LLM-P & 28.36 & 23.62 & 42.33 & 35.57 & 27.47 & 18.63 & 12.64 & 35.21 & 24.63 & 63.47 \\
 FILM & 5.97 & 5.97 & 11.17 & 11.17 & 16.55 & 0.49 & 0.49 & 4.84 & 4.84 & 33.68 \\
 Ours & 45.77 & 40.75 & 57.88 & 51.14 & 23.29 & 45.09 & 34.41 & 58.21 & 43.13 & 59.11 \\
 \midrule
  \multicolumn{11}{l}{\textbf{Target-specific Long}} \rowsqueeze\\
  \midrule
   LLM-P$^{*}$ & 5.97 & 5.14 & 18.91 & 17.26 & 60.56 & 1.52 & 0.82 & 15.28 & 13.05 & 78.03 \\
 LLM-P & 5.97 & 4.80 & 19.65 & 17.30 & 64.89 & 1.52 & 1.01 & 16.04 & 14.17 & 64.14 \\
 FILM & 0.00 & 0.00 & 4.14 & 4.14 & 79.17 & 0.00 & 0.00 & 6.26 & 6.26 & 70.14 \\
 Ours & 13.43 & 12.44 & 27.11 & 24.67 & 62.21 & 19.70 & 17.34 & 35.61 & 31.08 & 78.99 \\
 \midrule
  \multicolumn{11}{l}{\textbf{Abstract}} \rowsqueeze\\
 \midrule
 LLM-P$^{*}$ & 1.32 & 0.92 & 15.68 & 12.57 & 38.69 & 6.16 & 2.83 & 16.92 & 11.21 & 70.92 \\
 LLM-P & 3.95 & 2.33 & 16.78 & 12.45 & 36.27 & 6.16 & 3.58 & 18.15 & 12.42 & 67.20 \\
 FILM & 0.00 & 0.00 & 4.87 & 4.87 & 33.23 & 0.00 & 0.00 & 8.02 & 8.02 & 49.45 \\
 Ours & 10.53 & 8.09 & 24.23 & 19.68 & 35.90 & 9.59 & 5.74 & 21.30 & 15.01 & 61.54 \\
 \midrule
\end{tabular}
\label{table_1:procthor}
\vspace{-2mm}
\end{table*}

\begin{table*}[t]
\caption{Ablation experimental results of exploration strategies in the task-specific short setting, where No Exp. and No Front. represent no exploration and no frontiers exploration, respectively.}
\vspace{-1mm}
\centering
\footnotesize
\setlength{\tabcolsep}{12pt}
\begin{tabular}{@{}lcccccccccc@{}}
\toprule
\multirow{2}{*}{\textbf{Method}}& \multicolumn{5}{c}{\textbf{Normal-scale}}
& \multicolumn{5}{c}{\textbf{Large-scale}} \rowsqueeze \\
 \cmidrule{2-6}\cmidrule{6-11} \rowsqueeze
 & SR & PLWSR & GC & PLWGC & Path & SR & PLWSR & GC & PLWGC & Path \\
 \midrule
 No Exp. & 29.85 & 29.09 & 42.08 & 40.92 & 6.09 & 11.27 & 10.68 & 24.26 & 22.99 & 5.32 \\
 No Front. & 41.29 & 35.03 & 54.25 & 46.54 & 27.59 & 36.76 & 26.91 & 49.35 & 35.51 & 52.38 \\
 Ours & 45.77 & 40.75 & 57.88 & 51.14 & 23.29 & 45.09 & 34.41 & 58.21 & 43.13 & 59.11 \\
 \midrule

\end{tabular}
\label{table_1:ab_exp_s_and_l}
\vspace{-5mm}
\end{table*}

\subsection{Data Collection and Training}
\textbf{Data collection: }The training samples for the high-level planner consist of human instruction, current completed plans, current semantic feature maps and the groundtruth plan for the next step, and those for the low-level controller include plan for the next step, textual and semantic features for current border segments and the groundtruth action sequences representing primitives, location and targets.
We leverage GPT-4 and the ProcTHOR simulator to generate the large-scale dataset to train the LLaVA-based high-level planner. We annotate several seed instructions and leverage GPT-4 to generate more instructions and corresponding plans based on the object list for each scene in the ProcTHOR, where samples with logical errors are filtered with PDDL parameters \cite{shridhar2020alfred}. 
We then implement the generated plans in ProcTHOR and collect the navigation trajectories, RGB-D images, object locations and robot poses as the training data. Finally, the generated samples are parsed into high-level planning samples and low-level action data.

\textbf{Training: }We follow the supervised fine-tuning paradigm in LLM for training the LLaVA model in high-level planner and low-level controller, where we mask out $p_i$ and $\{a_j^i\}_{j=1}^{\tau_i}$ in the $i_{th}$ step. 
The CE loss leveraged in the training process is represented by:
\begin{align}
    \mathcal{L}_{\text{}}=-\mathbb{E}_{(\boldsymbol{X}_T,\boldsymbol{R})\sim\mathcal{D}}\Big[\sum_{m=1}^M\log p_{\boldsymbol{\theta}}(R_m|\boldsymbol{R}_{<m}, \boldsymbol{X}_{V},\boldsymbol{X}_T)\Big]
\end{align}
where $\boldsymbol{X}_{V}$ denotes scene feature maps and $\boldsymbol{X}_T$ means input text prompt tokens. $\boldsymbol{R}_{<m}$ represents the output text tokens before the $m_{th}$ token $R_m$ and $M$ are number of output tokens.
In the training stage, we propose to construct counterfactual samples to motivate the inference ability of the foundation model on EIF. Specifically, we remove the target objects in the scene descriptions from the original samples and replace them with target objects that have similar other properties such as usage through an artificial mapping method.

\section{Experiments}
\subsection{Implementation Details}
\textbf{Training configurations: }We employed the LLaVA-7B architecture with the Vincua-1.3-7B pre-training weights for the high-level planner and the low-level controller, which is finetuned with our generated data by the LoRA strategy. 
For the visual encoder, we sampled 32 visual embeddings from each frontier in the semantic feature maps up to 256 tokens as scene information representation. 
We generated 2k instructions with three subparts (1386 target-specific short, 333 target-specific long and 332 abstract instructions) for 2509 scenes in ProcTHOR, which results in 30k groundtruth plans for training the high-level planner. 
We implemented the plans in ProcTHOR with $A^{*}$ algorithm to collect the expert trajectory as the groundtruth for training low-level controller. 
Target-specific short and long instructions mean those containing objects to be interacted (e.g. Place the egg in the bowl) for task achievements, whose number of step plan is respectively lower than 15 and not. 
Abstract instructions do not contain the interacted objects in the instructions (e.g. Make a simple lunch for me). 
We also generate 201, 67 and 152 data for each subpart as the test set. 
We utilized 8 NVIDIA 3090 GPUs to finetune the high-level planner and the low-level controller for an hour in the training stage.

\textbf{Metrics: }
Following the ALFRED benchmark \cite{shridhar2020alfred}, we use success rates (SR), goal condition success (GC), path length and their path-length-weighted (PLW) counterparts for evaluation. 
SR means the ratio of the cases where the agent completely achieve the human instructions, and GC measures the ratio of objects in the state of goal achievements. 
PLWSR and PLWGC calculate SR and GC weighted by the expert trajectory planning step number divided by the actual execution step number, which measures the trade-off between performance and efficiency.
Path represents the distance moved by the robot, which is utilized to measure the efficiency of task planning.

\textbf{Simulated environments: }We perform extensive experiments in the ProcTHOR simulators, where the step size of translation and rotation for the agent is $0.25$m and $90^\circ$ respectively. ProcTHOR contains 10k house-level scenes with objects from 93 categories, where the agent receives $600 \times 600$ RGB-D images in the egocentric view. We divide the scenes into normal-scale ([0, 10]) and large-scale ([10, 16]) ones based on the side length of the room. 

\subsection{Comparison with Baselines}
Table \ref{table_1:procthor} demonstrates the results on ProcTHOR for LLM-Planner, FILM and our method, where our approach significantly outperforms the state-of-the-art-method LLM-Planner.
Although LLM-Planner utilizes the rich commonsense embedded in LLMs to generate plans for the agent, it fails to align the pre-trained LLMs with the scene information. The generated plans are usually infeasible due to the non-existence of the objects for interaction, and the re-planning module suffers from low success rate and low efficiency. On the contrary, our method construct the semantic feature maps which grounds the pre-trained multimodal LLMs to the realistic physical scene, and the unknown environment can be efficiently explored by understanding the visual clues for executable plan generation. 
In the target-specific short task setting, it is observed that our method outperforms LLM-Planner and FILM by 17.41\% and 39.80\% success rate in normal scale scenes, respectively. 
It is worth noting that our method loses less than 2\% success rate in transferring to large-scale scenes, while LLM-Planner and FILM lose 34\% and 91\% success rate, respectively, which demonstrates the excellent scalability of our method in scene scales.
Our approach remains leading in performance in more challenging target-specific long and abstract tasks.
Meanwhile, the leading PLWSR and PLWGC metrics verify that our low-level controller can find the target object at a lower navigation cost.
Moreover, the success rate of conventional methods (e.g., FILM) in the large-scale scenes is near zero, while our approach can achieve 9.59\% success rate. Since the service robot is usually deployed in house-level scenes, our method is proven to be more practical.

\begin{figure*}[t]
  \centering
    \includegraphics[width=1.0\textwidth]{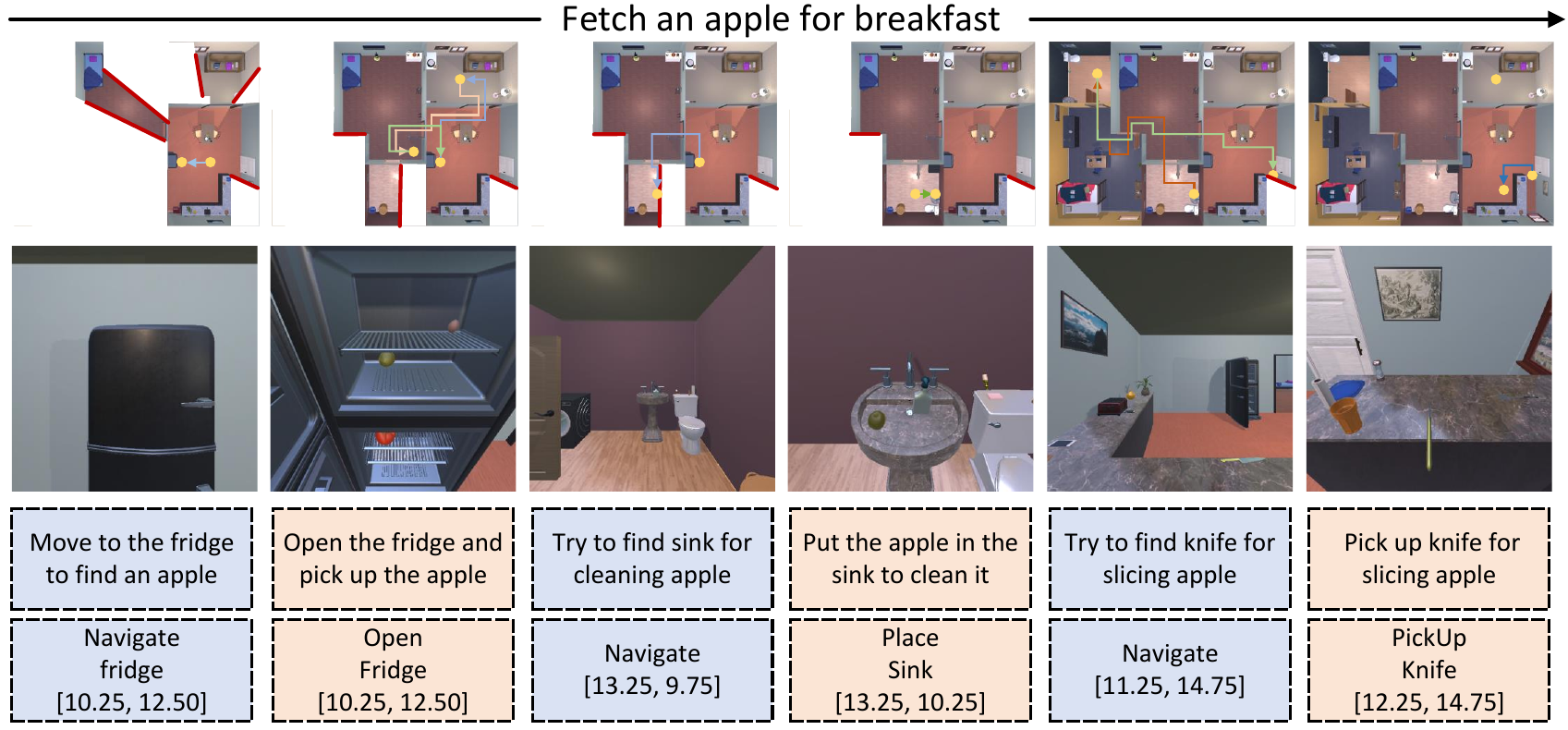}
  \caption{An example of EIF in unknown environments. The agent only navigates the task-related regions for visual clue collection with high efficiency, and generates feasible plans to complete the abstract instructions.}
  \label{fig:vis_example_1}
  \vspace{-5mm}
\end{figure*}

We demonstrate the qualitative results in Fig. \ref{fig:vis_example_1}, where we show the step-wise plan, the exploration process and the robot implementation during a whole sequence for EIF. 
In the beginning, the agent is initialized in the bedroom area and selects the navigation borders outside the room for exploration, as the instruction \textit{making breakfast} is irrelevant to bedrooms. 
During the navigation, the agent gradually knows to explore the kitchen area by observing the dining table and the counter, and it is even aware that opening the fridge may find food for breakfast due to the rich commonsense in our finetuned low-level controller. 
As a result, abstract instruction is achieved by serving diverse food for breakfast, where only related regions are navigated with high exploration efficiency in the unknown environment. Fig. \ref{fig:failure_case} illustrates the statistics of failure cases caused by different reasons. The failure mostly comes from unsuccessful navigation because of the large house-level scene, and the top reasons including \textit{too close to targets} and \textit{fail to see closed space} indicate that navigation algorithms should be designed with high compatibility of the subsequent manipulation.

\begin{table}[t]
\captionof{table}{Effectiveness of our generated plans and exploration actions.}
\vspace{-1mm}
\setlength{\tabcolsep}{4pt}
\footnotesize
\begin{tabular}{@{}lcccccccccc@{}}
\toprule
\multirow{2}{*}{\textbf{Method}} & & \multicolumn{2}{c}{\textbf{GT}} & \multicolumn{5}{c}{\textbf{Normal-scale \& target-specific short}} \rowsqueeze \\
\cmidrule{3-9} \rowsqueeze
& & Plan. & Exp. &SR & PLWSR & GC & PLWGC & Path \\
\midrule
\multirow{4}{*}{Ours} 
&& \textbf{$\checkmark$} & \textbf{$\checkmark$}  & 64.18 & 62.51 & 72.76 & 69.54 & 18.23 \\
&& \textbf{$\checkmark$} & -  & 49.75 & 47.20 & 60.07 & 56.38 & 21.64 \\
&& - & \textbf{$\checkmark$}  & 55.72 & 53.20 & 66.67 & 62.71 & 14.70 \\
&& - & -  & 45.77 & 40.75 & 57.88 & 51.14 & 23.29 \\
\midrule
\label{table:ab_gt}
\end{tabular}
\vspace{-7mm}
\end{table}

\begin{table}[t]
\centering
   \setlength{\tabcolsep}{4.5pt}
\footnotesize
\captionof{table}{Ablation study of different scene feature maps.}
\vspace{-1mm}
\begin{tabular}{@{}lcccccccc@{}}
\toprule
\multirow{2}{*}{\textbf{Method}} && \multicolumn{5}{c}{\textbf{Normal-scale \& target-specific short}} \rowsqueeze \\
\cmidrule{3-7} \rowsqueeze
&& SR & PLWSR & GC & PLWGC & Path \\
\midrule
No Feature && 36.36 & 29.53 & 45.73 & 38.06 & 16.46 \\
No Attention && 44.78 & 39.02 & 56.63 & 49.40 & 24.54 \\
Random Attention && 44.27 & 38.24 & 56.72 & 47.96 & 25.90 \\
Ours && 45.77 & 40.75 & 57.88 & 51.14 & 23.29 \\
\midrule
\label{table:ab_map}
\end{tabular}
\vspace{-1cm}
\end{table}

\subsection{Ablation Studies}
\begin{figure}[t]
 \centering
   \includegraphics[width=1.0\linewidth]{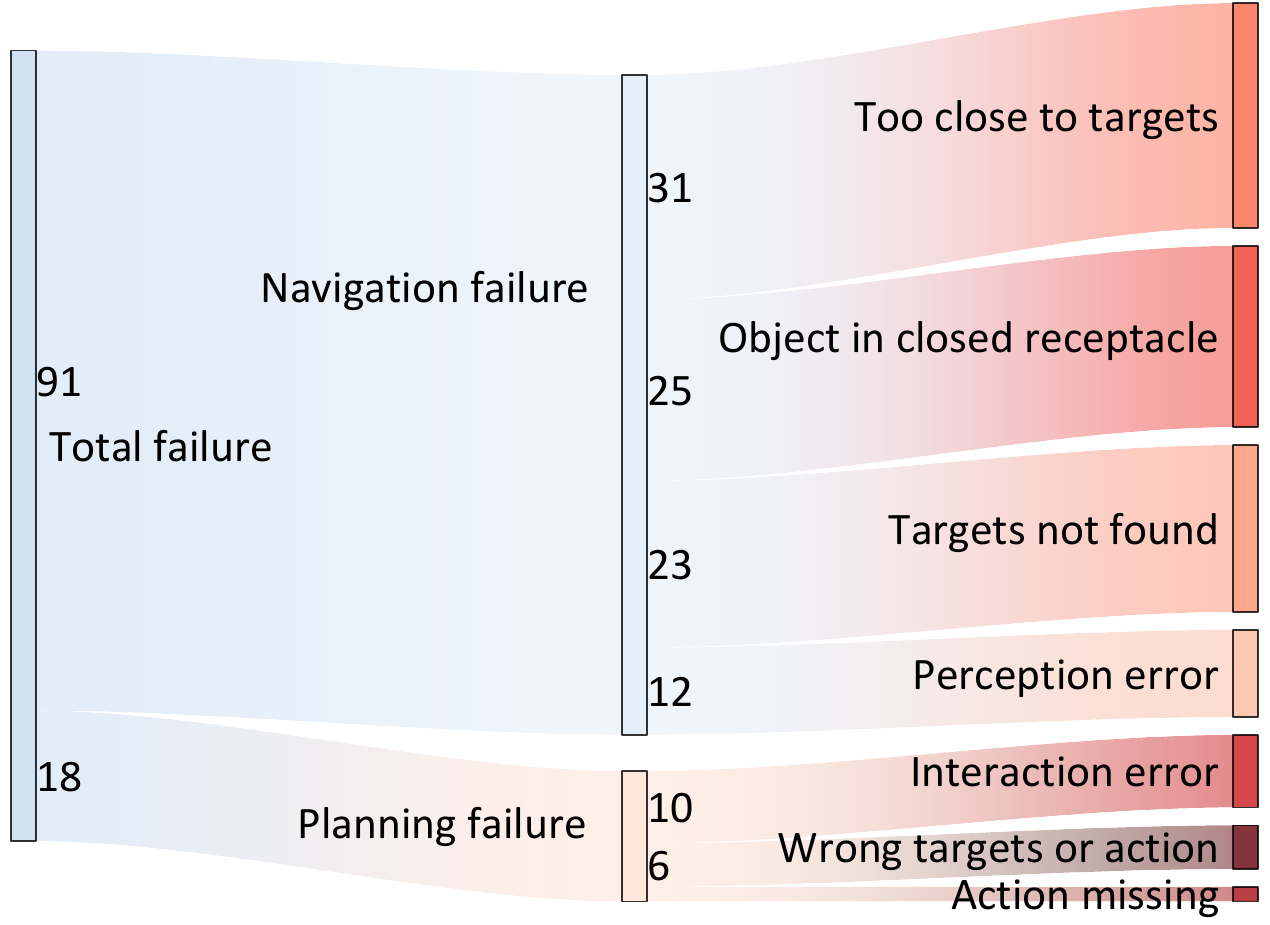}
    \caption{All failure cases on ProcTHOR simulator.}
    \label{fig:failure_case}
    \vspace{-8mm}
\end{figure}

\textbf{Effectiveness of the high-level planner and the low-level controller: }We evaluated the variants of our method where the planner and the controller are respectively replaced with the groundtruth step-wise plans and groundtruth action sequences.
It is important to note that some of the failure causes (e.g., too close to the target) illustrated in Fig. \ref{fig:failure_case} could not be resolved even with GT step-by-step planning and navigation goals.
Table \ref{table:ab_gt} demonstrates the results where the performance of our methods is close to that of the groundtruth, which indicates the effectiveness of our LLaVA-based planner and controller. 
Moreover, the performance of active exploration in low-level controller mainly influences the success rate, since it is important to find the correct objects to interact in unknown environments.
Meanwhile, low-level controller significantly impacts the path length since directly exploring the related regions enables the agent to accomplish the instruction faster.

\begin{figure}
    \centering
    \includegraphics[width=1.0\linewidth]{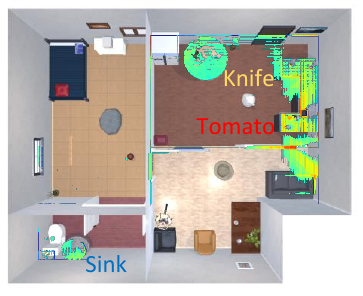}
    \caption{visualization of dynamic region attention weights.}
    \label{fig:attention_weight}
    \vspace{-7.5mm}
\end{figure}

\textbf{Effectiveness of the online semantic feature map: }The semantic feature map provides visual information of explored regions for the planner and the controller to generate feasible plans and efficient actions, and we report the performance of different semantic maps to validate the effectiveness of our method. 
Table \ref{table:ab_map} demonstrates the results for the settings of semantic maps with only category information, semantic feature maps without dynamic attention, semantic feature maps with random attention, and our semantic feature maps. The results demonstrate that the implicit rich semantic features are necessary for effective exploration of unknown environments, and the dynamic attention also enhances the performance of the semantic feature map as it removes the information redundancy for the large house-level scenes. 
We also visualize the dynamic region attention when the agent builds the semantic feature map in the unknown environment as illustrated in Fig. \ref{fig:attention_weight}. For the instruction \textit{Slice the tomato for salad}, the features of the kitchen area especially the tomato and the sink are considered with high attention(The green color represents greater weight), which indicates that the dynamic region attention learns relevant visual clues for feasible action generation.

\begin{table}[t]
\centering
\setlength{\tabcolsep}{4.5pt}
\footnotesize
\captionof{table}{Ablation experiment for high-level task planner.}
\vspace{-1mm}
\begin{tabular}{@{}lcccccc@{}}
\toprule
\multirow{2}{*}{\textbf{Method}} && \multicolumn{5}{c}{\textbf{Normal-scale \& target-specific short}} \rowsqueeze \\
\cmidrule{3-7} \rowsqueeze
&& SR & PLWSR & GC & PLWGC & Path \\
\midrule
FILM & & 5.97 &5.97  & 11.17 & 11.17 & 16.55 \\
Ours w/ BERT & & 24.38 & 18.38  & 39.81 &29.61 & 20.64 \\
Ours & & 45.77 & 40.75 & 57.88 & 51.14 & 23.29 \\
\midrule
\label{table:ab_exp_bert}
\vspace{-7mm}
\end{tabular}
\end{table}

\begin{table}[t]
\centering
\setlength{\tabcolsep}{4.5pt}
\footnotesize
\captionof{table}{Ablation experiment for foundation models.}
\vspace{-1mm}
\begin{tabular}{@{}lcccccc@{}}
\toprule
\multirow{2}{*}{\textbf{Method}} && \multicolumn{5}{c}{\textbf{Normal-scale \& target-specific short}} \rowsqueeze \\
\cmidrule{3-7} \rowsqueeze
&& SR & PLWSR & GC & PLWGC & Path \\
\midrule
GPT-4o & & 30.54 & 19.04 & 43.24 &27.34 & 20.11 \\
Conv-LLaVA-7B & & 44.33 & 26.31 & 53.62 &31.87 & 24.21 \\
LLaMA-Adapter-7B & & 43.84 &26.88  & 52.61 &32.64 & 25.34 \\
Ours & & 45.77 & 40.75 & 57.88 & 51.14 & 23.29 \\
\midrule
\label{table:ab_exp_mllm}
\vspace{-11mm}
\end{tabular}
\end{table}

\textbf{Effectiveness of active exploration:}
 Existing EIF frameworks often lack active exploration capabilities, making them difficult to deploy in unknown  environments. 
 Our approach addresses this limitation by utilizing pre-trained models to construct fine-grained semantic feature maps and leveraging foundation models to generate task planning and interaction actions based on these maps. 
 Table \ref{table_1:ab_exp_s_and_l} demonstrates the ablation experiments for different exploration strategies in the target-specific short setting.
 In house-level unknown environments, the no-exploration strategy reduces success rates by 15.92\% and 33.82\% for normal and large-scale settings, respectively, highlighting the importance of active exploration in unknown environment EIF tasks. 
 The efficiency of active frontier exploration is demonstrated by the fact that the success rate of the navigation strategy without frontier exploration is reduced by 4.48\% and 8.30\%, respectively, with comparable navigation costs compared to our approach.

\textbf{Influence w.r.t. high level planner:}
To further clarify the performance improvement of the model, we follow the FILM setting and use BERT to recognize the target objects from the instructions and generate high-level plans by filling the target objects into the corresponding parsing templates according to the predicted task categories.
Table \ref{table:ab_exp_bert} illustrates the results demonstrating that changing the LLaVA-7B to BERT occurred with performance decreases, and the performance still outperforms the FILM due to the ability of the low-level controller to explore unknown regions to find the target.


\textbf{Other foundation models:}
We further discuss the other foundation models and the results are illustrated in Table \ref{table:ab_exp_mllm}. 
GPT-4o achieved only a 30.54\% success rate, significantly lower than fine-tuned foundation models. Prompt-based grounding for downstream EIF tasks is ineffective compared to fine-tuning foundation models, indicating synthesized data requires post-processing before training. 

\section{Conclusion}
In this paper, we have proposed an EIF approach for unknown environments, where the agent is required to explore the environment efficiently to generate feasible action plans with existing objects to achieve human instructions. We first build a hierarchical EIF framework including a high-level planner and a low-level controller, and then build a semantic feature map with dynamic region attention to provide visual information for the planner and the controller. Extensive experiments demonstrate the effectiveness and efficiency of our framework in the house-level unknown environment. 
However, this work lacks real manipulation implementation and the designed navigation policy ignores the compatibility with manipulation. We will design mobile manipulation strategies for general tasks and implement the closed-loop system on real robots in the future.

\section*{Acknowledgments}
This work was supported in part by the National Natural Science Foundation of China under Grant 62376032, Grant 62125603 and A*STAR National Robotics Programme (RIE2025) M25N4N2009.

{\small
\bibliographystyle{ieee_fullname}
\bibliography{root}

\begin{thebibliography}{10}\itemsep=-1pt

\bibitem{batra2020objectnav}
Dhruv Batra, Aaron Gokaslan, Aniruddha Kembhavi, Oleksandr Maksymets, Roozbeh Mottaghi, Manolis Savva, Alexander Toshev, and Erik Wijmans.
\newblock Objectnav revisited: On evaluation of embodied agents navigating to objects.
\newblock {\em arXiv preprint arXiv:2006.13171}, 2020.

\bibitem{blukis2022persistent}
Valts Blukis, Chris Paxton, Dieter Fox, Animesh Garg, and Yoav Artzi.
\newblock A persistent spatial semantic representation for high-level natural language instruction execution.
\newblock In {\em CoRL}, pages 706--717. PMLR, 2022.

\bibitem{deitke2022️}
Matt Deitke, Eli VanderBilt, Alvaro Herrasti, Luca Weihs, Kiana Ehsani, Jordi Salvador, Winson Han, Eric Kolve, Aniruddha Kembhavi, and Roozbeh Mottaghi.
\newblock Procthor: Large-scale embodied ai using procedural generation.
\newblock {\em NeurIPS}, 35:5982--5994, 2022.

\bibitem{devlin2018bert}
Jacob Devlin, Ming-Wei Chang, Kenton Lee, and Kristina Toutanova.
\newblock Bert: Pre-training of deep bidirectional transformers for language understanding.
\newblock {\em arXiv preprint arXiv:1810.04805}, 2018.

\bibitem{ding2023embodied}
Mingyu Ding, Yan Xu, Zhenfang Chen, David~Daniel Cox, Ping Luo, Joshua~B Tenenbaum, and Chuang Gan.
\newblock Embodied concept learner: Self-supervised learning of concepts and mapping through instruction following.
\newblock In {\em CoRL}, pages 1743--1754. PMLR, 2023.

\bibitem{huang2023voxposer}
Wenlong Huang, Chen Wang, Ruohan Zhang, Yunzhu Li, Jiajun Wu, and Li Fei-Fei.
\newblock Voxposer: Composable 3d value maps for robotic manipulation with language models.
\newblock In {\em CoRL}, pages 540--562. PMLR, 2023.

\bibitem{jatavallabhula2023conceptfusion}
Krishna~Murthy Jatavallabhula, Alihusein Kuwajerwala, Qiao Gu, Mohd Omama, Tao Chen, Alaa Maalouf, Shuang Li, Ganesh Iyer, Soroush Saryazdi, Nikhil Keetha, et~al.
\newblock Conceptfusion: Open-set multimodal 3d mapping.
\newblock {\em arXiv preprint arXiv:2302.07241}, 2023.

\bibitem{kumar2024open}
Nishanth Kumar, William Shen, Fabio Ramos, Dieter Fox, Tom{\'a}s Lozano-P{\'e}rez, Leslie~Pack Kaelbling, and Caelan~Reed Garrett.
\newblock Open-world task and motion planning via vision-language model inferred constraints.
\newblock {\em arXiv preprint arXiv:2411.08253}, 2024.

\bibitem{mildenhall2021nerf}
Ben Mildenhall, Pratul~P Srinivasan, Matthew Tancik, Jonathan~T Barron, Ravi Ramamoorthi, and Ren Ng.
\newblock Nerf: Representing scenes as neural radiance fields for view synthesis.
\newblock {\em ACM}, 65(1):99--106, 2021.

\bibitem{min2021film}
So~Yeon Min, Devendra~Singh Chaplot, Pradeep~Kumar Ravikumar, Yonatan Bisk, and Ruslan Salakhutdinov.
\newblock Film: Following instructions in language with modular methods.
\newblock In {\em ICLR}, 2021.

\bibitem{mu2024embodiedgpt}
Yao Mu, Qinglong Zhang, Mengkang Hu, Wenhai Wang, Mingyu Ding, Jun Jin, Bin Wang, Jifeng Dai, Yu Qiao, and Ping Luo.
\newblock Embodiedgpt: Vision-language pre-training via embodied chain of thought.
\newblock {\em NIPS}, 36, 2024.

\bibitem{pashevich2021episodic}
Alexander Pashevich, Cordelia Schmid, and Chen Sun.
\newblock Episodic transformer for vision-and-language navigation.
\newblock In {\em ICCV}, pages 15942--15952, 2021.

\bibitem{rana2023sayplan}
Krishan Rana, Jesse Haviland, Sourav Garg, Jad Abou-Chakra, Ian Reid, and Niko Suenderhauf.
\newblock Sayplan: Grounding large language models using 3d scene graphs for scalable robot task planning.
\newblock In {\em CoRL}, 2023.

\bibitem{shridhar2020alfred}
Mohit Shridhar, Jesse Thomason, Daniel Gordon, Yonatan Bisk, Winson Han, Roozbeh Mottaghi, Luke Zettlemoyer, and Dieter Fox.
\newblock Alfred: A benchmark for interpreting grounded instructions for everyday tasks.
\newblock In {\em CVPR}, pages 10740--10749, 2020.

\bibitem{song2022one}
Chan~Hee Song, Jihyung Kil, Tai-Yu Pan, Brian~M Sadler, Wei-Lun Chao, and Yu Su.
\newblock One step at a time: Long-horizon vision-and-language navigation with milestones.
\newblock In {\em CVPR}, pages 15482--15491, 2022.

\bibitem{song2023llm}
Chan~Hee Song, Jiaman Wu, Clayton Washington, Brian~M Sadler, Wei-Lun Chao, and Yu Su.
\newblock Llm-planner: Few-shot grounded planning for embodied agents with large language models.
\newblock In {\em ICCV}, pages 2998--3009, 2023.

\bibitem{wu2023embodied}
Zhenyu Wu, Ziwei Wang, Xiuwei Xu, Jiwen Lu, and Haibin Yan.
\newblock Embodied task planning with large language models.
\newblock {\em arXiv preprint arXiv:2307.01848}, 2023.

\bibitem{yang2023lacma}
Cheng-Fu Yang, Yen-Chun Chen, Jianwei Yang, Xiyang Dai, Lu Yuan, Yu-Chiang~Frank Wang, and Kai-Wei Chang.
\newblock Lacma: Language-aligning contrastive learning with meta-actions for embodied instruction following.
\newblock {\em arXiv preprint arXiv:2310.12344}, 2023.

\bibitem{yin2024sg}
Hang Yin, Xiuwei Xu, Zhenyu Wu, Jie Zhou, and Jiwen Lu.
\newblock Sg-nav: Online 3d scene graph prompting for llm-based zero-shot object navigation.
\newblock {\em NeurIPS}, 37:5285--5307, 2024.

\bibitem{yin2025unigoal}
Hang Yin, Xiuwei Xu, Linqing Zhao, Ziwei Wang, Jie Zhou, and Jiwen Lu.
\newblock Unigoal: Towards universal zero-shot goal-oriented navigation.
\newblock In {\em CVPR}, pages 19057--19066, 2025.

\bibitem{yu2023l3mvn}
Bangguo Yu, Hamidreza Kasaei, and Ming Cao.
\newblock L3mvn: Leveraging large language models for visual target navigation.
\newblock In {\em IROS}, pages 3554--3560. IEEE, 2023.

\end{thebibliography}
}

\end{document}